\documentclass[3p,times,procedia]{elsarticle}
\usepackage{ecrc}
\usepackage[bookmarks=false]{hyperref}
    \hypersetup{colorlinks,
      linkcolor=blue,
      citecolor=blue,
      urlcolor=blue}
\volume{00}

\firstpage{1}

\journalname{Procedia Computer Science}

\runauth{Areeg Fahad, M. Zarkoosh, Safa F., Sana Sabah}

\jid{procs}
\usepackage{graphicx} % Required for inserting images
\usepackage{url}
\usepackage{hyperref}
\usepackage{amssymb}
\usepackage[table]{xcolor}
\definecolor{lightgreen}{rgb}{0.88, 1, 0.88}
\definecolor{lightorange}{rgb}{1, 0.93, 0.8}
\definecolor{lightred}{rgb}{1, 0.8, 0.8}
\usepackage[figuresright]{rotating}
\begin{document}
\begin{frontmatter}

%% Title, authors and addresses

%% use the tnoteref command within \title for footnotes;
%% use the tnotetext command for the associated footnote;
%% use the fnref command within \author or \address for footnotes;
%% use the fntext command for the associated footnote;
%% use the corref command within \author for corresponding author footnotes;
%% use the cortext command for the associated footnote;
%% use the ead command for the email address,
%% and the form \ead[url] for the home page:
%%
%% \title{Title\tnoteref{label1}}
%% \tnotetext[label1]{}
%% \author{Name\corref{cor1}\fnref{label2}}
%% \ead{email address}
%% \ead[url]{home page}
%% \fntext[label2]{}
%% \cortext[cor1]{}
%% \address{Address\fnref{label3}}
%% \fntext[label3]{}

%\dochead{International Conference on Machine Learning and Data Engineering}%%%
%% Use \dochead if there is an article header, e.g. \dochead{Short communication}
%% \dochead can also be used to include a conference title, if directed by the editors
%% e.g. \dochead{17th International Conference on Dynamical Processes in Excited States of Solids}

%\title{TaskComplexity: Developing a Dataset for Programming Task Classification}
\title{TaskComplexity: A Dataset for Task Complexity Classification  with In-Context Learning, FLAN-T5 and GPT-4o Benchmarks}

%% use optional labels to link authors explicitly to addresses:
%% \author[label1,label2]{<author name>}
%% \address[label1]{<address>}
%% \address[label2]{<address>}

\author{Areeg Fahad Rasheed\corref{cor1}} 
\author{M. Zarkoosh}
\author{Safa F. Abbas}
\author{Sana Sabah Al-Azzawi}
% \author[a,b]{Third Author}

\address[a]{College of Information Engineering, Al-Nahrain University, Baghdad, Iraq}
\address[b]{Software Engineering, Baghdad, Iraq}
\address[c]{Computer Techniques Engineering Department, Al-Rasheed University College, Baghdad, Iraq}
\address[d]{EISLAB Machine Learning, Lule University of Technology, Luleå, Sweden}

\begin{abstract}

 This paper addresses the challenge of classifying and assigning programming tasks to experts, a process that typically requires significant effort, time, and cost. To tackle this issue, a novel dataset\footnote{https://github.com/AREEG94FAHAD/TaskComplexityEval-24} containing a total of 4,112 programming tasks was created by extracting tasks from various websites. Web scraping techniques were employed to collect this dataset of programming problems systematically. Specific HTML tags were tracked to extract key elements of each issue, including the title, problem description, input/output, examples, problem class, and complexity score. Examples from the dataset are provided in the appendix to illustrate the variety and complexity of tasks included. The dataset's effectiveness has been evaluated and benchmarked using two approaches; the first approach involved fine-tuning the FLAN-T5 small model on the dataset, while the second approach used in-context learning (ICL) with the GPT-4o mini.  The performance was assessed using standard metrics: accuracy, recall, precision, and F1-score. The results indicated that in-context learning with GPT-4o-mini outperformed the FLAN-T5 model.

\end{abstract}

\begin{keyword}
 GPT-4o-mini, Flan-T5, task classification, in-context learning, Natural Language Processing (NLP), dataset creation.

%% keywords here, in the form: keyword \sep keyword

%% PACS codes here, in the form: \PACS code \sep code

%% MSC codes here, in the form: \MSC code \sep code
%% or \MSC[2008] code \sep code (2000 is the default)

\end{keyword}
\cortext[cor1]{Corresponding author. ORCID.: 0000-0002-7078-6200}
\end{frontmatter}

%\correspondingauthor[*]{Corresponding author. Tel.: +0-000-000-0000 ; fax: +0-000-000-0000.}
\email{areeg.fahad@coie-nahrain.edu.iq, fahedareeg@gmail.com}

%%
%% Start line numbering here if you want
%%
% \linenumbers
\vspace*{-6pt}
%% main text

%\enlargethispage{-7mm}
\section{Introduction}
\label{main}

In recent years, artificial intelligence has integrated into our lives and is used in various aspects such as healthcare, image recognition \cite{rasheed2024enhancing, rasheed2023multi}, image generation, and other applications \cite{talib2023deep, alnedawe2023new, jasim2020ecg, wang2024scgan}. Natural language processing (NLP) is a type of AI specialized in the analysis of continuous text and finding relationships within it \cite{yao2024survey, rasheed-zarkoosh-2024-mashee}. NLP is used in different tasks, such as text classification, text generation, question answering, and more \cite{rasheed2023arabic, tan2023can, sindhu2024evolution,sabry2022hat5,nilsson2022leveraging,al-azzawi-etal-2023-nlp}. 

Programming skills are considered crucial and are recognized as among the essential skills required by many companies \cite{shute1991likely}. The classification of programming tasks typically demands significant effort and can be costly. Since no existing dataset for classifying such tasks was available, programming and logical tasks were collected from different websites, including Kattis \cite{kattis}, LeetCode \cite{leetcode}, HackerRank \cite{hackerrank}, and Topcoder \cite{topcoder}, among others. Web scraping \cite{khder2021web, singrodia2019review} was employed in the collection process. A total of 4,112 tasks were compiled in the dataset. To provide a more comprehensive understanding of the types of tasks included, examples from the dataset are provided in the appendix.
Two different machine-learning approaches were utilized for the classification process. The first approach involved the fine-tuning of the FLAN-T5 model \cite{chung2024scaling}, and the second approach utilized in-context learning \cite{rasheed-zarkoosh-2024-mashee} with the GPT-4o-mini model \cite{sonoda2024diagnostic}. 
The following paragraph highlights Why such a dataset is important.

%\vspace{0.5cm}
\begin{itemize}
    \item Efficient Resource Allocation: The dataset can be leveraged to develop an efficient system for automatic resource allocation. This includes assigning tasks to individuals based on their skill levels and optimizing the use of computational resources according to the complexity of the problems.
    
    \item Recommendation Systems: This dataset can be used to create a tool for recommending programming challenges based on a user's past performance and preferences on online programming training websites. Such a system can enhance the user experience and promote continuous learning.

    \item Teacher Support: When teaching programming languages, teachers often encounter challenges in selecting exam questions that adequately cover the diverse skill levels of all students. This dataset can be used to create a model that assists in building balanced exams, ensuring comprehensive coverage, and addressing varying skill levels.
    
    \item Task Complexity Prediction: The dataset can be used to create a platform that predicts the complexity of new tasks 
\end{itemize}

 \textbf {The paper's key contribution} is the creation and evaluation of a novel dataset comprising 4,112 programming tasks extracted from various websites using web scraping techniques. This dataset addresses the challenge of efficiently classifying and assigning programming tasks to experts, potentially revolutionizing resource allocation, recommendation systems, and educational support in the field of programming. 
 
 The rest of the paper is organized as follows: Section 2 explains the procedure used for dataset collection. Section 3 illustrates the two approaches used in the evaluation process. Section 4 presents the results of each approach and discusses the main findings. Finally, Section 5 provides the main conclusions and offers suggestions for future work.

\section{Dataset Collection Process}
\begin{figure}
    \centering
    \includegraphics[width=1\linewidth]{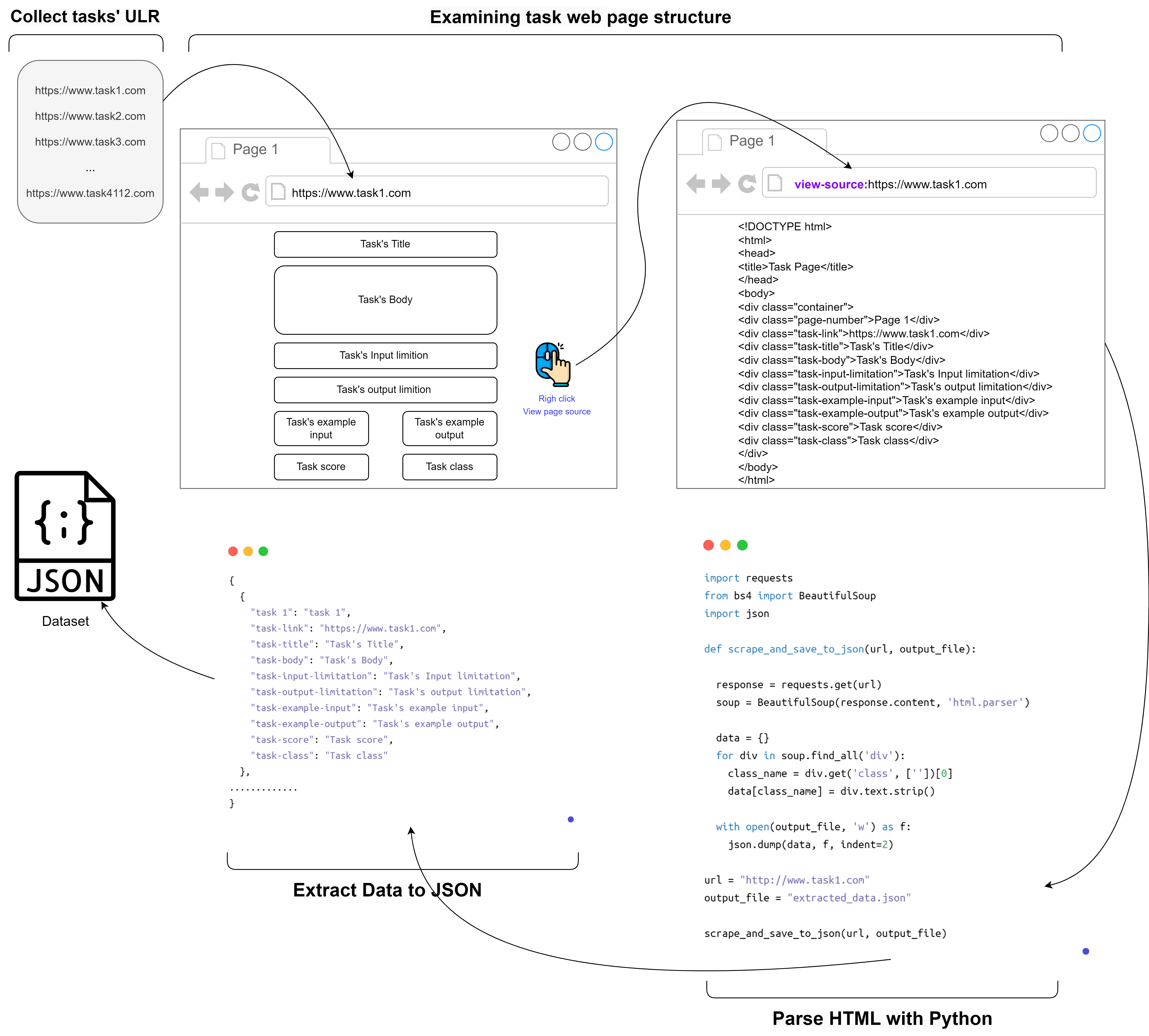}
    \caption{Dataset collection process}
    \label{fig:dig}
\end{figure}

A range of tasks for training and developing programming and logical skills is offered by many websites, including platforms such as Kattis \cite{kattis}, LeetCode \cite{leetcode}, HackerRank \cite{hackerrank}, and Topcoder \cite{topcoder}, among others. The dataset collection process is illustrated in Figure \ref{fig:dig}. Initially, 4,112 links to programming task pages were manually gathered. The HTML structure of these pages was then examined to determine their accessibility for scraping. Task details, including the title, task description, input limitations, output limitations, example inputs, example outputs, task class, and complexity score, were extracted using web scraping. These details were saved as JSON objects in a JSON file. It is important to note that some websites prevent access to their HTML content, which limits the ability to scrape data from those sources. The available tasks were categorized into three classes: simple, medium, and hard, with complexity scores ranging from 1 to 9.7.

\section{Methodology }

This study employed two distinct machine learning approaches for the classification process: fine-tuning a large language model based on Flan-T5 and in-context learning using the GPT-4o-mini model. The following subsections provide a detailed description of each approach. The first subsection gives an overview of the Flan-T5 fine-tuning process, followed by a description of the in-context learning methodology using GPT-4o-mini.
\subsection{Flant-T5 Fine-tuning }

FLAN-T5 is a Large Language Model (LLM) created in 2022 by Google; it is a modified version of T5 and is trained on a wide range of datasets \cite{chung2024scaling}. Flan-T5 is used as an LLM for many purposes, such as text generation, text classification, and question answering, as well as for other translations \cite{woisetschlager2024federated}. Flan-T5 is open-source and freely available. There are different versions of Flan-T5: the first one is called Small with 77 million parameters, the Base with 248 million parameters, the Large with 783 million parameters, the X-Large with 2.85 billion parameters, and the final one is the XX-Large with 11.3 billion parameters. Increasing the number of parameters can enhance the model's capabilities, but it requires more computational resources \cite{mei2024efficiency}. Due to the lack of access to computational resources, the small version of the Flan-T5 was used. 
The settings for the Flan-T5 fine-tuning are presented in the table \ref{tab:props}. 

\begin{figure}
    \centering
    \includegraphics[width=0.5\linewidth]{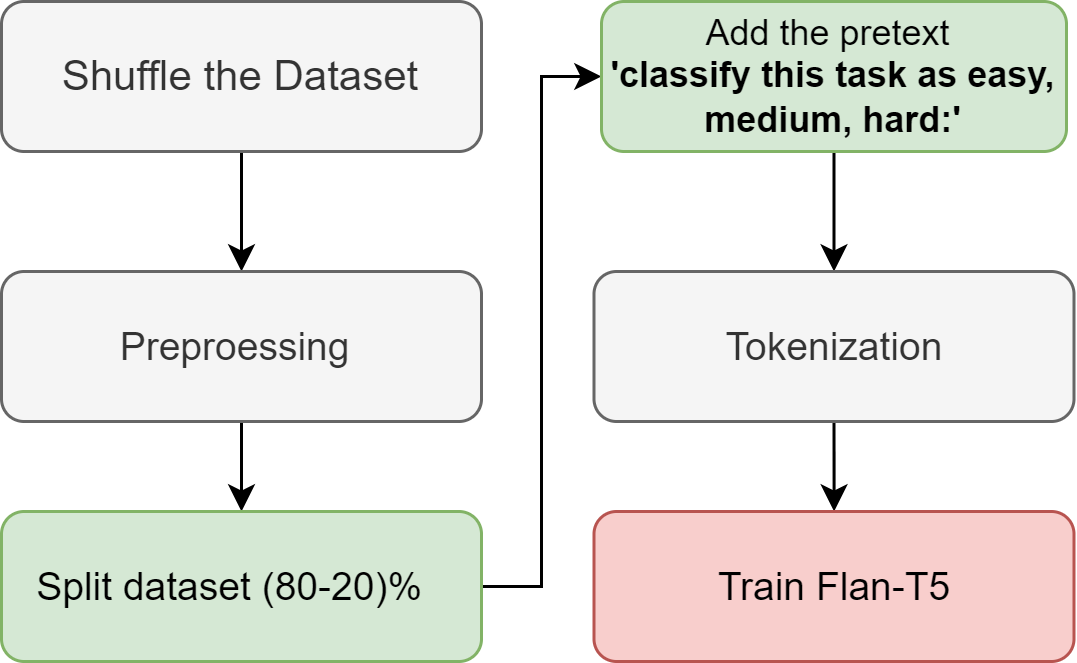}
    \caption{Finetuning FLAN-T5 steps}
    \label{fig:flant5dig}
\end{figure}

The fine-tuning process for FLAN-T5 followed a structured approach, as shown in Figure \ref{fig:flant5dig}. The first step involves shuffling the dataset. Following this, the preprocessing stage was implemented to convert the data into the proper format to input into the model. The preprocessed dataset was then divided into training and testing subsets, with an 80-20\% split ratio, respectively.  Each sample within the dataset was subsequently tokenized, with a maximum token length of 1300. The tokenized data is fed to the model for training. It is important to note that the pretext 'classify this task as easy, medium, hard:' is added to each sample for better performance.

\begin{table}[h]
    \centering
    \begin{tabular}{cc}
    \hline
        Property & Value \\ \hline
        Model Version & Flan-T5-small \\ \hline
        Number of Classes & 3 \\ \hline
        Training Samples & 3289 \\ \hline
        Test Samples & 823 \\ \hline
        Max Tokens & 1300 \\ \hline
        Learning Rate & 5e-4 \\ \hline
        Train and Test Batch Size & 8 \\ \hline
        Seed & 42 \\ \hline
    \end{tabular}
    \caption{Flan-T5 fine-tuning settings}
    \label{tab:props}
\end{table}

\subsection{In context learning based on GPT-4o-mini}

In-context learning is a powerful approach utilized in large language models (LLMs) that enables task performance without the need for fine-tuning.

In this framework, the model is trained on a wide range of datasets and then used to perform other tasks without fine-tuning \cite{trad2024prompt, sahoo2024systematic}.
When employing ICL, instead of further training the model for specific tasks, task-relevant examples are provided within the input prompt. The model then leverages these examples to infer how to perform the requested task, effectively "learning" from the context provided.

This method has several advantages: it can be used when there is a lack of available datasets or to eliminate the computational requirements of the fine-tuning process. Generally, there are three types of in-context learning: zero-shot learning, where the model is directly asked to perform the task without prior examples; one-shot learning, where the pre-trained model is given a single instance before performing the task; and few-shot learning, where the model is provided with a few examples \cite{ma2024fairness, li2024promptad}.

GPT-4o-mini is OpenAI's latest cost-efficient small model. It is designed to expand AI applications by offering affordable intelligence. With a context window of 128K tokens and multimodal support, it is utilized for various tasks, including classification, text generation, question answering, summarization, and more.

\begin{figure}
    \centering
    \includegraphics[width=0.9975\linewidth]{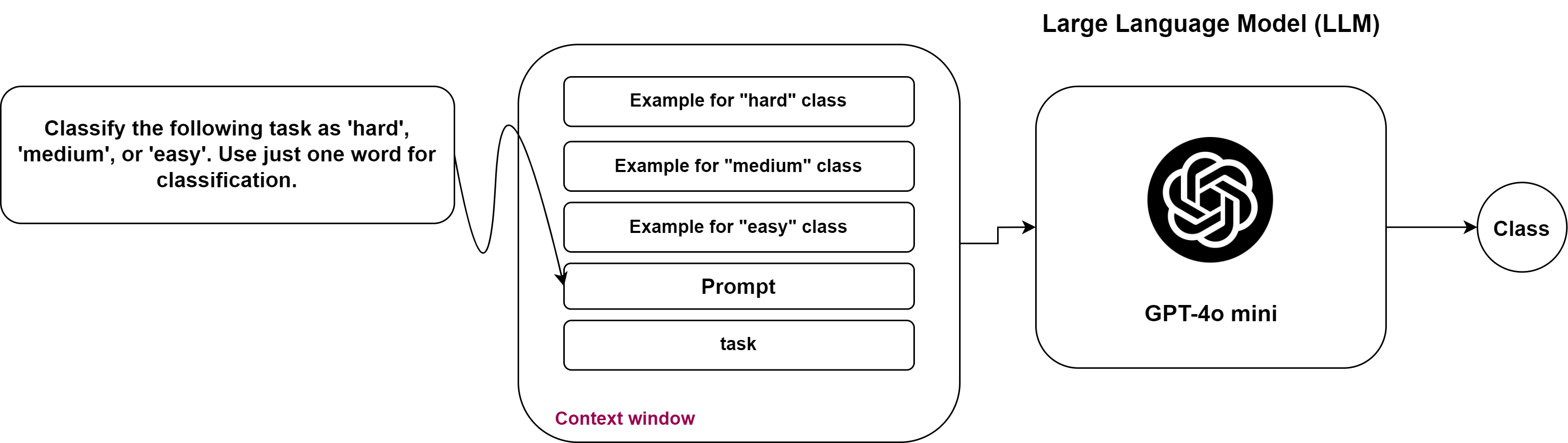}
    \caption{In context learning based on GPT-4o-mini}
    \label{fig:chatgpt4mindig}
\end{figure}

Figure \ref{fig:chatgpt4mindig} illustrates how ICL is applied using GPT-4o-mini; the process begins by randomly selecting three samples from the dataset, one for each class (e.g., easy, medium, hard). These samples are used as few-shot examples, providing the model with context. These examples are placed at the top of the model’s context window.

Next, a prompt is appended below the examples: "Classify the following task as 'hard', 'medium', or 'easy'. Use just one word for classification." The task that needs classification is placed after the prompt within the context window.

Once the entire context window (including the few-shot examples, the prompt, and the task) is prepared, it is sent via the OpenAI API for inference using GPT-4o-mini. The model processes the input and returns a classification result (either 'hard', 'medium', or 'easy'). This result is then saved for each sample and used for future evaluation of the model’s performance.

\section{Results and discussion}
\begin{figure}
    \centering
    \includegraphics[width=0.75\linewidth]{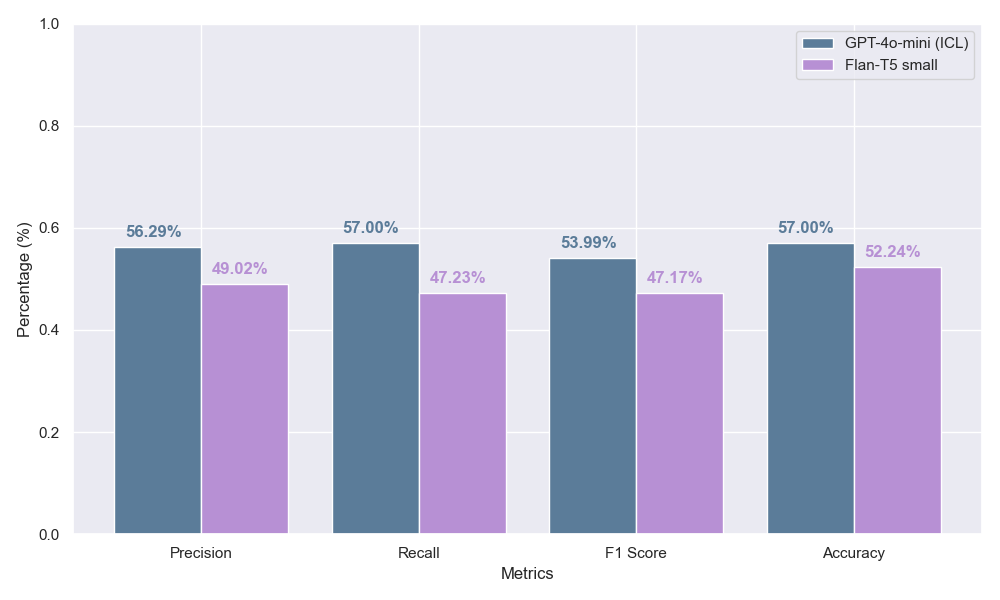}
    \caption{Performance Comparison: In context learning based on GPT-4o-mini vs. Flan-T5 small}
    \label{fig:results}
\end{figure}

The model's performance was evaluated using different metrics such as accuracy, F1-score, precision, and recall. The results, illustrated in Figure \ref{fig:results}, show that the in-context learning approach based on GPT-4o-mini outperforms the fine-tuned Flan-T5 small version across all metrics. 
Specifically, GPT-4o-mini achieved 57.00\% accuracy compared to 52.24\% for Flan-T5. For the F1-score, GPT-4o-mini achieved 53.99\%, while Flan-T5 reached 47.17\%. In terms of recall, GPT-4o-mini achieved 57.00\%, whereas Flan-T5 achieved 47.23\%. Lastly, in precision, GPT-4o-mini reached 56.29\%, surpassing the 49.02\% achieved by Flan-T5.

Overall, using in-context learning with GPT-4o-mini or fine-tuning the Flan-T5 small version does not show high performance for classifying the complexity of the programming tasks. The main reason is that in traditional task classification, samples belonging to the same class often share common properties, such as specific keywords or phrases, which help the model identify and learn from these patterns during training. However, this is not the case with our dataset. Each sample, even if it pertains to the same class, exhibits different structures and meanings. This variability hinders the GPT-4o-mini and FLAN-T5 small version model's ability to recognize and utilize consistent patterns for classification purposes.

\section{Conclusion and Future suggestion}

This paper introduces a new dataset for classification and predicting the complexity of programming tasks. The dataset is collected from various platforms offering programming logic tasks using web scraping techniques. The proposed dataset contains a total of 4,112 programming tasks and is organized in a JSON file format. The dataset includes properties such as task name, task description, input description and limitations, output description and limitations, example input and output, along with problem class and complexity. 

The dataset is evaluated using different approaches. The first approach is fine-tuning, where we used the Flan-T5 small version. The second approach is in-context learning, utilizing GPT-4o-mini, the computationally efficient version of ChatGPT.

The results indicate that in-context learning performs better than Flan-T5, even without fine-tuning the dataset. However, the overall results suggest that the dataset requires a more capable model since each problem has its own structure, with different input and output limitations, as well as unique examples. It becomes challenging for the model to identify patterns across problems within the same class, as GPT-4o-mini and Flan-T5 achieved accuracies of only 57% % and 52%, respectively.

For future work, we suggest using other more capable versions of FLAN-T5, such as Large, X-Large, and XX-Large. Additionally, we are planning to explore other LLMs such as LLaMA-3 \cite{dubey2024llama, touvron2023llama}, and ChatGPT-4 \cite{wu2023brief}. We are also interested in collecting more samples to create a larger dataset and improve classification performance.

\bibliographystyle{elsarticle-num}
\bibliography{PROCS_ICMLDE2024}

\newpage
\section{Appendix}
In this section, we provide three examples from our dataset for each class.

\textit{An example of an easy problem }
\\
%\subsection*{An example of easy problem }
\rowcolors{1}{}{lightgreen}  % Alternating row colors

\begin{tabular}{|>{\columncolor{lightgreen}}m{4cm}|>{\columncolor{lightgreen}}m{10cm}|}
    \hline
    \textbf{Title} & \textbf{A Shortcut to What?} \\ 
    \hline
    \textbf{Problem Description} & Write a program that reads an integer $n$ from the input. The program should add five to $n$, then triple the value and finally subtract ten. You are welcome to simplify the formula if you can, as long as the program gives the correct results. \\ 
    \hline
    \textbf{Input} & Input consists of one line containing one integer $n$, where $-1000 \leq n \leq 1000$. \\ 
    \hline
    \textbf{Output} & Output consists of one line containing one integer, the value computed from $n$ as described above. \\ 
    \hline
    \textbf{Sample Input} & -4 \\ 
    \hline
    \textbf{Sample Output} & -7 \\ 
    \hline
    \textbf{Problem Class} & Easy \\ 
    \hline
    \textbf{Complexity Score} & 1.1 \\ 
    \hline
    \textbf{URL} & https://open.kattis.com/problems/shortcuttowhat \\ 
    \hline
\end{tabular}
\\
\\

\newpage

\noindent
%\subsection*{An example of medium problem}
\textit{An example of medium problem }
\\

\rowcolors{1}{lightorange}{lightorange} % Apply light orange to all rows
\begin{tabular}{|m{4cm}|m{10cm}|}
    \hline
    \textbf{Title} & \textbf{Money Matters} \\ 
    \hline
    \textbf{Problem Description} & 
    Our sad tale begins with a tight clique of friends. Together they went on a trip to the picturesque country of Molvania. During their stay, various events which are too horrible to mention occurred. The net result was that the last evening of the trip ended with a momentous exchange of "I never want to see you again!"s. A quick calculation tells you it may have been said almost 50 million times! 
    
    Back home in Scandinavia, our group of ex-friends realize that they haven’t split the costs incurred during the trip evenly. Some people may be out several thousand crowns. Settling the debts turns out to be a bit more problematic than it ought to be, as many in the group no longer wish to speak to one another, and even less to give each other money.

    Naturally, you want to help out, so you ask each person to tell you how much money she owes or is owed, and whom she is still friends with. Given this information, you’re sure you can figure out if it’s possible for everyone to get even, and with money only being given between persons who are still friends.\\
    \hline
    \textbf{Input} &
    The first line contains two integers, $n$ ($2 \leq n \leq 10\,000$) and $m$ ($0 \leq m \leq 50\,000$), the number of friends and the number of remaining friendships. The friends are named $0, 1, \ldots, n-1$. Then $n$ lines follow, each containing an integer $o$ ($-10\,000 \leq o \leq 10\,000$) indicating how much each person owes (or is owed if $o < 0$). The first of those lines gives the balance of person $0$, the second line the balance of person $1$, and so on. The sum of these values is zero. After this comes $m$ lines giving the remaining friendships, each line containing two integers $x$, $y$ ($0 \leq x < y \leq n-1$), indicating that persons $x$ and $y$ are still friends. \\
    \hline
    \textbf{Output} & 
    Your output should consist of a single line saying \texttt{POSSIBLE} or \texttt{IMPOSSIBLE}. \\
    \hline
    \textbf{Sample Input} & 
    \begin{verbatim}
    5 3
    100
    -75
    -25
    -42
    42
    0 1
    1 2
    3 4
    \end{verbatim} \\
    \hline
    \textbf{Sample Output} & \texttt{POSSIBLE} \\
    \hline
    \textbf{Problem Class} & Medium \\
    \hline
    \textbf{Complexity Score} & 2.9 \\
    \hline
    \textbf{URL} & https://open.kattis.com/contests/pxw8se/problems/moneymatters\\ 
    \hline
\end{tabular}

%%%%%%%%%%%%%%%%%%%%%%%%%%%

%%%%%%%%%%%%%%%%%%%%%%%%%%%%

\subsection*{An example of hard problem}

\rowcolors{1}{lightred}{lightred} % Apply near-red color to all rows
\begin{tabular}{|m{4cm}|m{10cm}|}
    \hline
    \textbf{Title} & \textbf{Tug of War} \\ 
    \hline
    \textbf{Problem Description} & 
    A tug of war is to be arranged at the local office picnic. For the tug of war, the picnickers must be divided into two teams. Each person must be on one team or the other; the number of people on the two teams must not differ by more than 1; the total weight of the people on each team should be as nearly equal as possible. \\
    \hline
    \textbf{Input} & 
    The first line of input contains $n$, the number of people at the picnic. $n$ lines follow. The first line gives the weight of person 1; the second the weight of person 2; and so on. Each weight is an integer between 1 and 450. There is at least 1 and at most 100 people at the picnic. \\
    \hline
    \textbf{Output} & 
    Your output will be a single line containing 2 integers: the total weight of the people on one team, and the total weight of the people on the other team. If these numbers differ, give the lesser first. \\
    \hline
    \textbf{Sample Input} & 
    \begin{verbatim}
    3
    100
    90
    200
    \end{verbatim} \\
    \hline
    \textbf{Sample Output} & 190 200 \\
    \hline
    \textbf{Problem Class} & Hard \\
    \hline
    \textbf{Complexity Score} & 7.1 \\
    \hline
    \textbf{URL}& https://open.kattis.com/contests/m2gwtg/problems/tugofwar
\end{tabular}

\end{document}